\documentclass[conference]{IEEEtran}
\IEEEoverridecommandlockouts

\usepackage{cite}
\usepackage{amsmath,amssymb,amsfonts}
\usepackage{algorithmic}
\usepackage{graphicx}
\usepackage{textcomp}
\usepackage{xcolor}
\usepackage{bm}
\usepackage{amsmath}
\usepackage{amssymb}
\usepackage{enumitem}
\usepackage{multirow}
\usepackage{multicol}
\usepackage{floatrow}

\def\BibTeX{{\rm B\kern-.05em{\sc i\kern-.025em b}\kern-.08em
    T\kern-.1667em\lower.7ex\hbox{E}\kern-.125emX}}
\begin{document}

\title{Deep Technology Tracing for High-tech Companies}

\newcommand{\zk}[1]{\textcolor{red}{~Z:[#1]}}

\author{Han Wu, Kun Zhang, Guangyi Lv, Qi Liu, Runlong Yu, Weihao Zhao, Enhong Chen\IEEEauthorrefmark{1}, Jianhui Ma \\Anhui Province Key Laboratory of Big Data Analysis and Application \\School of Computer Science and Technology, University of Science and Technology of China \\ \{wuhanhan, zhkun, gylv, yrunl, zhaoweihao\}@mail.ustc.edu.cn, \{cheneh, qiliuql, jianhui\}@ustc.edu.cn}

\maketitle
\vspace{-0.2cm}
\let\thefootnote\relax\footnotetext{$^{*}$ denotes the corresponding author}

\begin{abstract}

	Technological change and innovation are vitally important, especially for high-tech companies. However, factors influencing their future research and development (R\&D) trends are both complicated and various, leading it a quite difficult task to make technology tracing for high-tech companies. To this end, in this paper, we develop a novel data-driven solution, i.e., \textit{Deep Technology Forecasting (DTF) framework}, to automatically find the most possible technology directions customized to each high-tech company. Specially, DTF consists of three components: Potential Competitor Recognition (PCR), Collaborative Technology Recognition (CTR), and Deep Technology Tracing (DTT) neural network. For one thing, PCR and CTR aim to capture competitive relations among enterprises and collaborative relations among technologies, respectively. For another, DTT is designed for modeling dynamic interactions between companies and technologies with the above relations involved. Finally, we evaluate our DTF framework on real-world patent data, and the experimental results clearly prove that DTF can precisely help to prospect future technology emphasis of companies by exploiting hybrid factors. 

\end{abstract}


\section{Introduction}
\label{s:introduction}

Technological change and innovation are important factors for productivity and competitiveness~\cite{kurtossy2004innovation}, especially for high-tech companies whose lifelines depend much on research and development (R\&D) achievements.
However, R\&D processes are often time and labor consuming and the available funds are usually limited~\cite{park2017application}. 
Therefore, there is a great need to develop efficient technology management techniques for high-tech companies~\cite{ernst2003patent}, so that they can make accurate demand estimates, apply fairness resource allocations, enhance innovation ability, and thus create competitive advantages in the fierce market circumstances.

In view of the importance of technology management, many efforts have been made in this area, including technology prospecting~\cite{ernst2003patent, kim2017novel, erdi2013prediction}, R\&D portfolio value analysis~\cite{park2017application}, competitor monitoring~\cite{ernst2003patent}, and so on. In particular, technology forecasting aims to measure the innovation degree of technologies and prospect their success possibility in the future, which are often based on quantitative analysis with indicators~\cite{kim2017novel, ernst2003patent} or holistic analyses of technologies in the whole market place~\cite{erdi2013prediction}. 
Few of them can be customized to each company's personalized needs as well as their dynamic evolving trends. For this reason, we try to find a possible solution by forecasting the emerging technologies suitable for each high-tech company automatically, to provide some data-driven insights on their future R\&D directions.

Indeed, there are many domain and technological challenges inherent in designing effective solutions to this problem. First, factors influencing future R\&D trends of companies are both complicated and various, including the effect of internal and external factors~\cite{del1999resource}, i.e., their own technical strengths and weaknesses and technological trend in the whole market place. Second, there exist many complex relations: 1) In order to survive from the fierce competition, companies often keep sensitive to the R\&D tendency of their competitors, i.e., competitive relations; 2) Some technologies are usually closely related and show a bundled synchronization, i.e., collaborative relations. Both of them have potential effects on firms' R\&D strategies, while can not be easily captured and modeled. Third, no matter technologies or company themselves are continuous to evolve, so another challenge is how to model dynamic interactions between companies and technologies and capture their potential evolving trends. 

To conquer the above challenges, in this paper, we propose a novel \textit{Deep Technology Forecasting (DTF) framework} to automatically identify the most emerging technologies that a company tends to develop further. Specially, DTF consists of three components: Potential Competitor Recognition (PCR), Collaborative Technology Recognition (CTR), and Deep Technology Tracing (DTT) neural network. For one thing, PCR and CTR aim to capture competitive relations among enterprises and collaborative relations among technologies, respectively. 
For another, DTT is introduced for modeling the dynamic interactions between companies and technologies with the above relations involved. 
Finally, extensive experiments are conducted on real-world patent data, whose results prove that DTF can precisely prospect future technology directions customized to given companies by exploiting hybrid factors.

\section{Data Description}
\label{s:data_description}
In this section, we first describe the public patent data we use, and then provide some supportive statistics.

\subsection{Data Description}
Patenting is one of the most important ways to protect core business concepts and proprietary technologies~\cite{liu2018patent}. 
Therefore, most of high-tech companies keep filing patents every year to protect their products, services and ideas.  
Since 1972, more than 6 million patent documents have been issued and granted in the United States Patent and Trademark Office (USPTO)\footnote{https://www.uspto.gov}, and number of patent assignees has reached 389,246, where more than 89\% are companies or corporations. 
So to speak, patents provide us with an open window for analyzing technology evolution of high-tech companies. 

\begin{figure}[h]
	\vspace{-0.1cm}
	\centering\includegraphics[height = 0.8in]{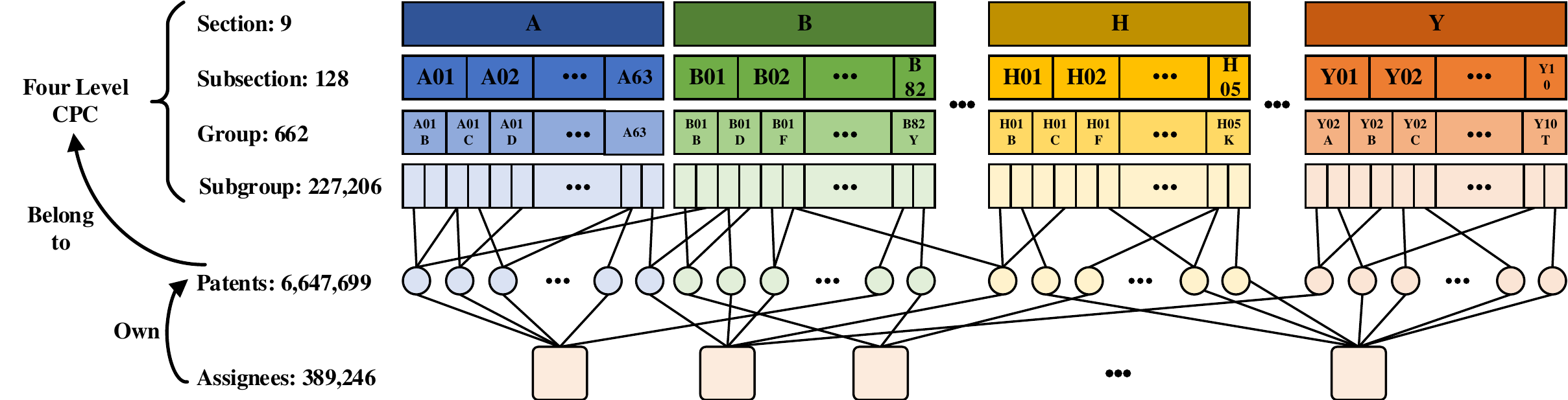}
	\caption{The visualization of Cooperative Patent Classification (CPC).} \label{fig:cpc}
	\vspace{-0.1cm}
\end{figure}
\vspace{-0.2cm}

In order to map patent pieces to technologies, we utilize the widely used Cooperative Patent Classification (CPC)\footnote{https://en.wikipedia.org/wiki/Cooperative\_Patent\_Classification}.
In fact, CPC is a patent classification system, which has been jointly developed by the European Patent Office (EPO) and the USPTO. As shown in Fig.~\ref{fig:cpc}, CPC has four levels. From the top down, technology categories are partitioned more and more detailed. 
For example, the first level 'section' has 9 classifications, and the code 'H' represents 'Electricity'; the third level has 662 classifications and 'H04J' means 'Multiplex Communication'. In general, each US patent is allocated several CPC codes according to their involved technologies at the beginning of its application. Therefore, given a company, we can find all its applied or granted patents as well as their corresponding technologies represented by CPC codes. 

\subsection{Statistics on Companies and Technologies}
In this part, we give some data statistics for revealing several supportive observations of companies and technologies. 

\begin{figure}[h]
	\vspace{-0.2cm}
	\centering\includegraphics[height = 1.0in]{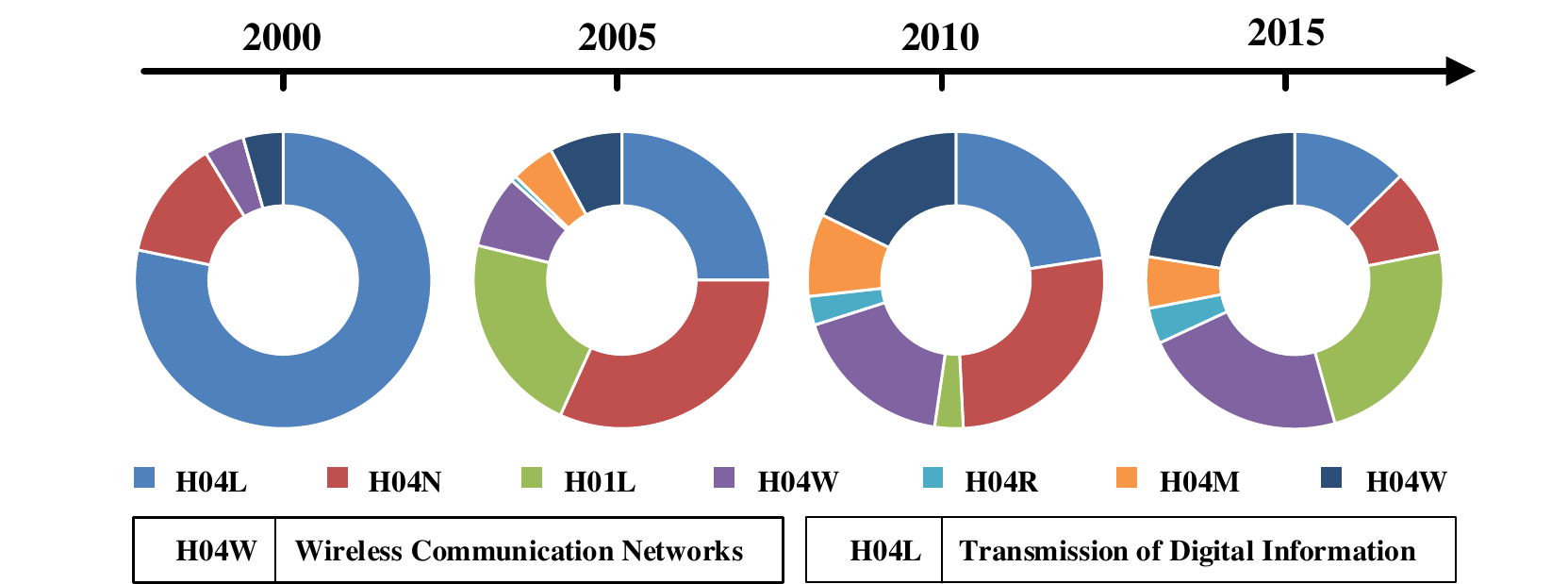}
	\caption{The technology evolving  trend of Apple Inc. from 2000 to 2015.} \label{fig:apple}\vspace{-0.1cm}
\end{figure}
\vspace{-0.1cm}

Fig.~\ref{fig:apple} depicts the evolving trend of 7 typical technologies of Apple Inc. from 2000 to 2015. Here we can see it is continually changing with time: some technologies keep increasing while some decreasing, i.e., the growing 'H04W' and the shrinking 'H04L'. It may tell the development trend that 'H04W' acts as Apple's current technology emphasizes and may potentially keep increasing in the next few years.

Then, we analyze the technology distribution (based on CPC section) of different types of companies shown in Fig.~\ref{fig:distribution}, from which we have three observations: 

\begin{figure}[h]
	\vspace{-0.1cm}
	\centering\includegraphics[height = 1.45in]{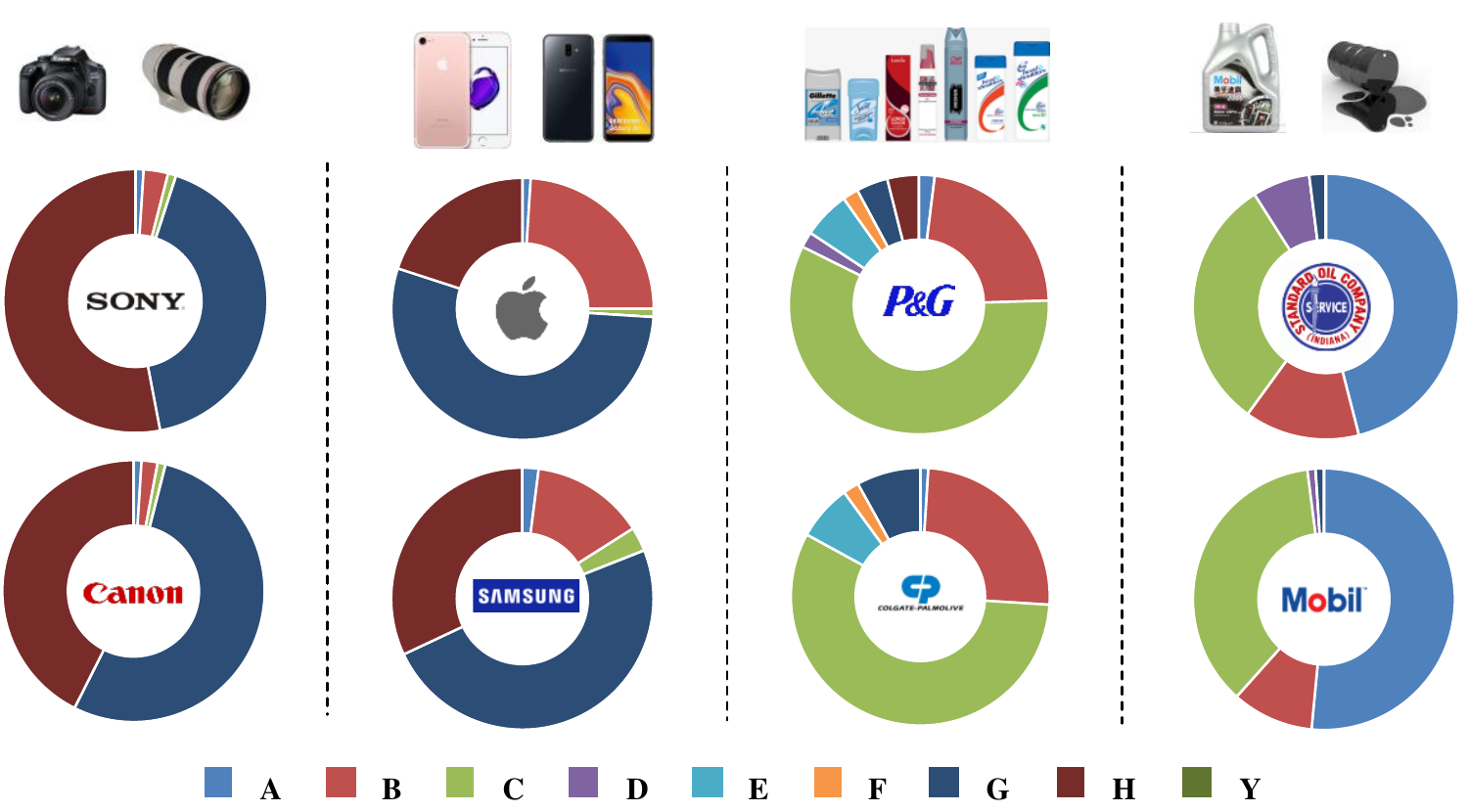}
	\caption{Technology distribution of different types of companies.} \label{fig:distribution}\vspace{-0.1cm}
\end{figure}
\vspace{-0.1cm}

\begin{itemize}[leftmargin=*,itemsep=2.5pt]\setlength{\itemsep}{0pt}
	\item Each company has its own technical strengths and weaknesses, indicated by the varying proportions of different technologies. For example, the Procter \& Gamble Company shows a great advantage in technology 'C' (Chemistry), while a disadvantage in technology 'G' (Physics).
	\item Companies who tend to be competitors share similar technology distributions. For instance, the top 3 technology categories of both Apple and Samsung are 'G' (Physics), 'H' (Electricity) and 'B' (Performing Operations; Transporting).
	\item Technology distributions among different types of companies vary a lot, which can be easily found from any two columns of Fig.~\ref{fig:distribution}.
\end{itemize}

\begin{figure}[h]
	\centering\includegraphics[height = 1in]{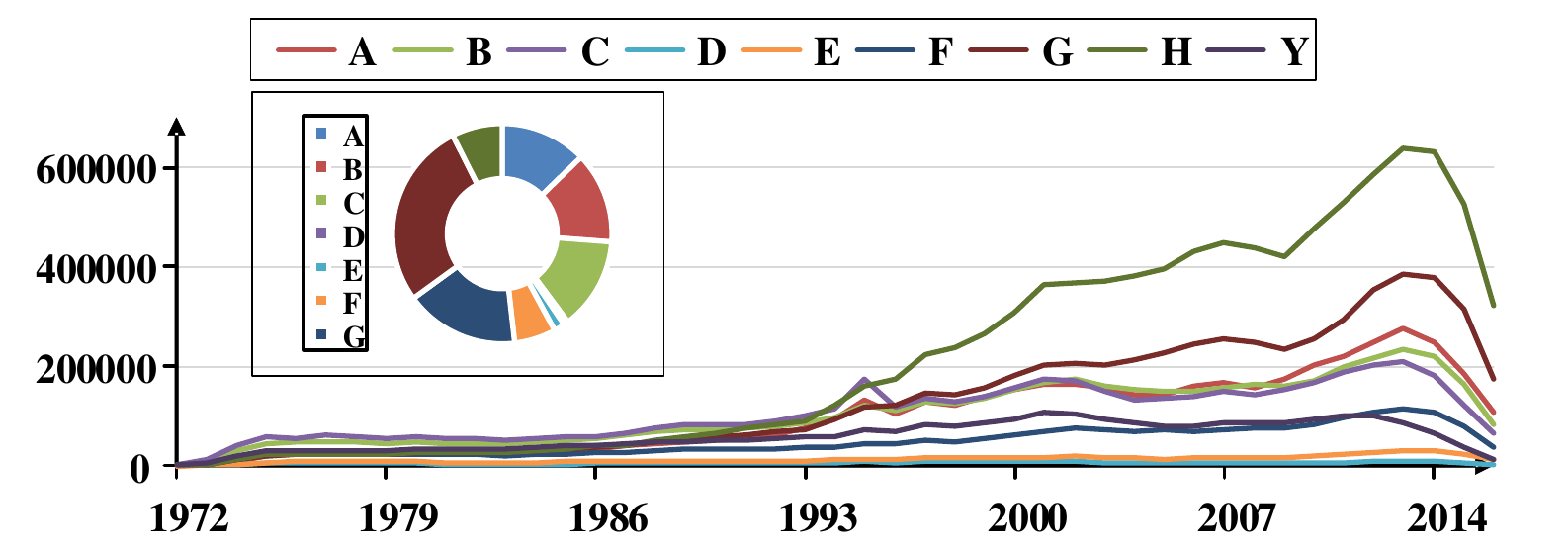}
	\caption{Technology evolving trend from 1972 to 2016 based on CPC section.} \label{fig:technology-trend}
	\vspace{-0.1cm}
\end{figure}

Fig.~\ref{fig:technology-trend} shows the number of patents granted in different CPC sections from 1972 to 2016. Here we can see a booming increase of most technologies, and the growth of some technologies seems kind of synchronous. For example, section 'H' (Electricity) and 'G' (Physics) have a very similar trend, which might benefit by the rapid development of the information industry, especially electronic hardwares like semiconductors. 

\vspace{-0.1cm}
\vspace{-0.1cm}

\begin{figure*}[t] 
	\centering\includegraphics[height = 2.15in]{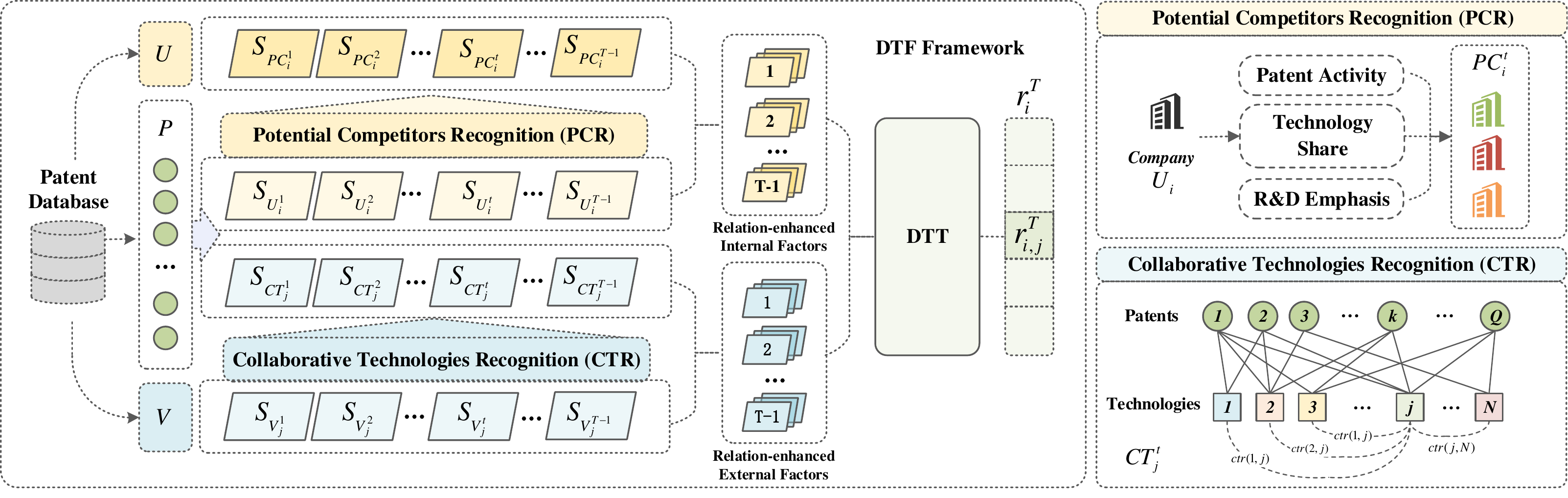}
	\caption{Overall Architecture of Deep Technology Forecasting (DTF) Framework.} 
	\label{fig:framework}\vspace{-0.2cm}
\end{figure*}

The above interesting observations can be instructive and meaningful, from which we can summarize following instructions for predicting the R\&D directions of a given company:

\begin{itemize}[leftmargin=*,itemsep=2.5pt]\setlength{\itemsep}{0pt}
	\item \textit{Internal and external factors}. When predicting the R\&D directions of a given company, we need to consider both internal factors, i.e., its original technical strengths and weaknesses, and external factors, i.e., the development trend of technologies in overall market place.
	\item \textit{Relations among companies and technologies}. Competitive relations among companies and collaborative relations among technologies can be also a great help.
	\item \textit{Dynamics of companies and technologies}. Both the companies and technologies keep evolving consistently, so we need also to model the dynamic interactions among them.
\end{itemize}
\section{Problem Statement}
Suppose there are $\small M$ companies ($\small{U={U_i|i=1,2,\cdots,M}}$), $\small N$ technologies ($\small V={V_j|j=1,2,\cdots,N}$) and $\small Q$ patents ($\small P={P_k|k=1,2,\cdots,Q}$) within $T$ years in patent database. Then, for company $\small U_i \in U$, its patent filing history can be represented by $\small{S_{U_i}=[S_{U_i^1},S_{U_i^2},\cdots,S_{U_i^t},\cdots,S_{U_i^T}]}$, where $\small S_{U_i^t}$ indicates the set of patents that $\small U_i$ files in year $t$. Similarly, for technology $\small V_j\in V$, its patent filing records can also be denoted as $\small{S_{V_j}=[S_{V_j^1},S_{V_j^2},\cdots,S_{V_j^t},\cdots,S_{V_j^T}]}$, where $\small S_{V_j^t}$ indicates the set of patents filed in year $t$ belonging to $\small V_j$.  Specifically, technology distribution of $\small U_i \in U$ in year $t$ is defined as:
	\begin{equation}
	    \label{equ:technology_distritbition}
	    \begin{array}{l}
	    {\small r_i^t=[r_{i,1}^t, r_{i,2}^t,\cdots, r_{i,j}^t,\cdots,r_{i,N}^t]}, \\
	    {\small r_{i,j}^t=\frac{|S_{U_i^t}\cap S_{V_j^t}|}{|S_{U_i^t}|}}, \\
	    \end{array}
    \end{equation}
    \noindent where $\small{|S_{U^t_i}\cap S_{V^t_j}|}$ means the number of patents belonging to $\small V_j$ that $\small U_i$ files in year $t$. Obviously, 
    if $U_i$ files a large number of patents belonging to $\small V_j$ in year $t$, we will have a big $\small r_{i,j}^t$, indicating that $\small U_i$ pays a great emphasis on $\small V_j$ in year $t$. 
    
Then we can formalize our research problem as follows: Given the patent filing history of a company $\small U_i$ before year $\small T$, $\small{S_{U_i}=[S_{U^1_i},S_{U^2_i},\cdots,S_{U^{T-1}_i}]}$, and that of a technology $\small V_j$, $\small{S_{V_j}=[S_{V^1_j},S_{V^2_j},\cdots,S_{V^{T-1}_j}]}$, our goal is to
predict $r_{i,j}^T$, and thus the whole technology distribution of $\small U_i$ in year $\small T$, represented by $\small{r_i^T=[r_{i,1}^T, r_{i,2}^T,\cdots,r_{i,N}^T]}$.

\begin{figure*}
	\centering\includegraphics[height = 1.7in]{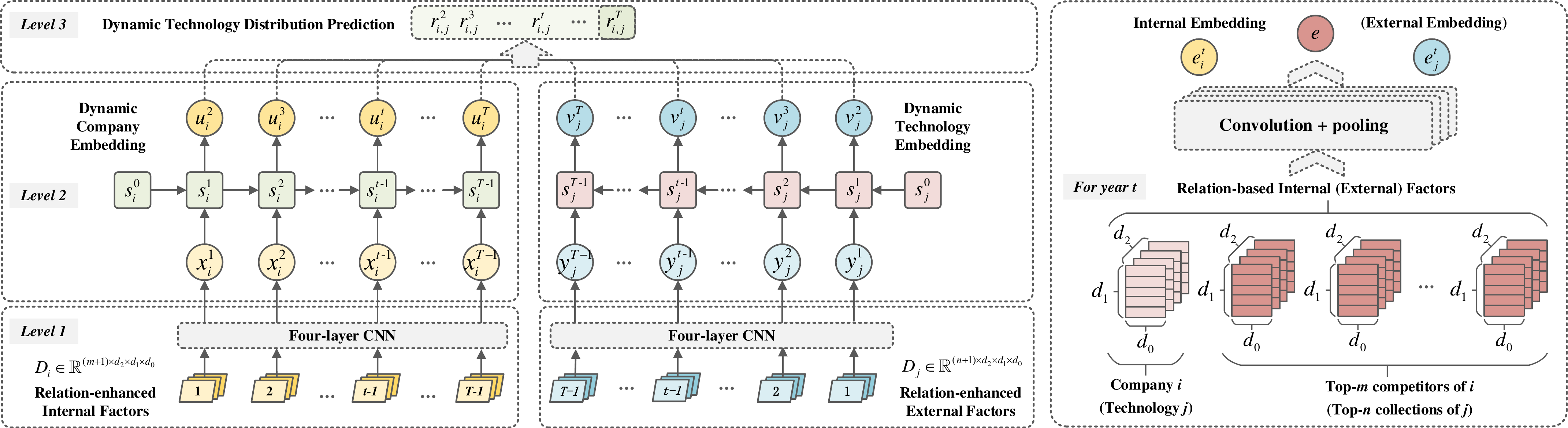}
	\caption{The architecture of Deep Technology Tracing (DTT) Neural Network.} \label{fig:RNN}
\end{figure*}

\section{\emph{DTF} Framework}
\label{s:model}

In this section, we provide a possible solution to the technology tracing problem, i.e., \emph{Deep Technology Forecasting (DTF)} framework shown in Fig.~\ref{fig:framework}, including Potential Competitors Recognition (PCR), Collaborative Technology Recognition (CTR), and Deep Technology Tracing (DTT) neural network.

\subsection{Potential Competitors Recognition (PCR)} 

Given a company $\small U_i \in U$, PCR aims to find its most likely competitors $\small PC_i^t \subset U$ in year $t$. Inspired by~\cite{ernst2003patent}, we apply three commonly used patent indicators for evaluating competitions among companies:

\begin{itemize}[leftmargin=*,itemsep=2.5pt]
	\item \textit{Patent Activity} ($\small I_1 = |S_{U_i^t}\cap S_{V_j^t}|$) is a fundamental patenting indicator. Decreasing or increasing of $I_1$ can be interpreted as changing levels of R\&D activity, and therefore, future technological and commercial~\cite{ernst2003patent}. 
	
	\item \textit{Technology Share} ($\small I_2 = {|S_{U_i^t}\cap S_{V_j^t}|}/{|S_{V_j^t}|}$) is based on patent applications, which measures a firm's competitive position in a technological field.
	
	\item \textit{R\&D Emphasis} ($\small I_3 = {|S_{U_i^t}\cap S_{V_j^t}|}/{|S_{U_i^t}|}$) illustrates the importance placed on a specific technological field within a firm's entire R\&D portfolio.
\end{itemize}

Then, we develop a competitive score for measuring competitive degrees based on the commonly used Euclidean distance. Specially, for $\small U_{i_1}\in U$ and $\small U_{i_2}\in U$ in year $t$, the competitive degree between them are denoted as: 
\begin{equation}
\label{equ:competition_score}
\small pcr^t(U_{i_1},U_{i_2}) = \sqrt{\sum_{q=1}^{3} \alpha_q(I_q^{U_{i_1},t}-I_q^{U_{i_2},t})^2}, \\
\end{equation}
\noindent where $\small I_q^{U_{i_1},t}$, $\small I_q^{U_{i_2},t}$ represents the $q$th indicator of $\small U_{i_1}$ and $\small U_{i_2}$ in year $t$ respectively, and $\alpha_q$ is the corresponding weight of $\small I_q$. Through Eq.\ref{equ:competition_score}, given $\small U_i \in U$ in year $t$, we can rank and get its top-$m$ potential competitors, indicated by $\small PC_i^t$. 

\subsection{Collaborative Technology Recognition (CTR)}
As shown in Fig.~\ref{fig:framework}, for each year, we first construct a bipartite whose nodes are patents and technologies while edges represent the ownership between them. In detail, if $\small P_k\in P$ belongs to $\small V_j\in V$, there will be an edge connecting $\small P_k$ and $\small V_j$. Then, a weighted network can be established, whose nodes are technologies and edges are their collaborations. Here, weight of edge between $\small V_{j_1}$ and $\small V_{j_2}$ is calculated by:
\begin{equation}
\label{equ:ctr}
\small ctr^t(V_{j_1}, V_{j_2}) = {|S_{V_{j_1}^t}\cap S_{V_{j_2}^t}|}/{|S_{V_{j_1}^t}\cup S_{V_{j_2}^t}|},\\
\end{equation}
\noindent where $\small{|S_{V_{j_1}^t}\cap S_{V_{j_2}^t}|}$ means the number of common patents shared by $\small V_{j_1}$ and $\small V_{j_2}$ in year $t$, and $\small{|S_{V_{j_1}^t}\cup S_{V_{j_2}^t}|}$ represents the total number of patents filed in $\small V_{j_1}$ and $\small V_{j_2}$ in year $t$. Naturally, bigger $\small ctr^t(V_{j_1}, V_{j_2})$ indicates a deeper collaboration. In this way, given $\small V_j \in V$ in year $t$, we can rank and get its top-$n$ collaborations, indicated by $\small CT_j^t$.

\subsection{Deep Technology Tracing (DTT) Neural Network} 

Fig.~\ref{fig:RNN} shows architecture of Deep Technology Tracing (DTT) Neural Network, which can be partitioned into three levels: \textit{1)} relation-enhanced factor representation; \textit{2)} dynamic embedding for companies and technologies; \textit{3)} final prediction for a given company and technology.

\subsubsection{\textbf{Relation-enhanced Factor Representation}}

As the first level of DTT, this part aims at learning the semantic representation of relation-enhanced internal and external factors. 

As shown in the right part of Fig.~\ref{fig:RNN}, each patent is combined with a sequence of words $e=[e_1, {e}_2, \cdots, {e}_{d1}]$, where ${e}_i\in \mathbb{R}^{d_0}$ is initialized by $d_0$-dimensional pre-trained word embedding and $d_1$ is the length of ${e}$. 
Then, for each company in each year, we totally sample $d_2$ patents as its internal factors. Then, patents of one company can be depicted by a tensor ${D}\in \mathbb{R}^{d_2*d_1*d_0}$. With the top-$m$ competitors extracted by PCR, we totally get $m+1$ company tensors in each year. In this way, relation-enhanced internal factors of company $\small U_i$ can be represented by $\small {D}_{i}\in \mathbb{R}^{(m+1)*d_2*d_1*d_0}$. Similar operations are applied in external factor extraction, so we also have $\small {D}_{j}\in \mathbb{R}^{(n+1)*d_2*d_1*d_0}$, i.e. the relation-enhanced external factor tensor in each year.

Next, we try to transform the above $\small {D}_{i}$ and $\small {D}_{j}$ into lower semantic embeddings through the commonly used convolutional neural network (CNN)~\cite{goldberg2016primer}. Three layers of convolution-pooling processes are set to gradually summarize the global interactions of words in a patent and finally reach a vectorial representation one $\dot{{e}}\in \mathbb{R}^d$, where $d$ is the output dimension of one patent document. Thus, company $i$ who have $d_2$ patents in each year can be represented as ${a}_i=\sigma({\dot{{e}}_1,\dot{{e}}_2,\cdots,\dot{{e}}_{d_2}})$, where ${a}_i\in \mathbb{R}^d$ and $\sigma$ is a mean value function. Along this line, the relation based internal factor tensor $\small {D}_{i}\in \mathbb{R}^{(m+1)*d_2*d_1*d_0}$ can be transformed into $\small {D}_{i}\in \mathbb{R}^{(m+1)*d}$. 

So, the relation-enhanced internal factor embedding of company $\small U_i$ in year $t$ is given by Eq.~(\ref{equ:internal}), where $\small pcr^t(U_i,U_{i'})$ is the competition score calculated in PCR, and ${a}_{i}^t\in \mathbb{R}^d$ is the patent embedding of $\small U_i$ in year $t$.
\begin{equation}
\label{equ:internal}
\small {x}_i^t={a}_{i}^t + \sum_{i'\in PC_i^t} pcr^t(U_i,U_{i'})*{a}_{i'}^t.\\
\end{equation}

Similarly, the relation-enhanced external factor embedding of technology $j$ in year $t$ is given by Eq.~(\ref{equ:external}), where $\small ctr^t(V_j,V_{j'})$ is the collaborative score calculated in CTR, and ${a}_{j}^t\in \mathbb{R}^d$ is the patent latent embedding of $\small V_j$ in year $t$. 
\begin{equation}
\label{equ:external}
\small {y}_j^t={a}_{j}^t + \sum_{j'\in CT_j^t} ctr^t(V_j,V_{j'})*{a}_{j'}^t.\\
\end{equation}

\subsubsection{\textbf{Dynamic Embedding for Companies \& Technologies}}
We employ Gated Recurrent Unit (GRU)~\cite{cho2014properties} to model the dynamic interactions of companies and technologies. As depicted in Fig.~\ref{fig:RNN}, given the yearly internal factor embedding sequence of company $\small U_i$, i.e., $\small{{x}_i=\{{x}_i^1,{x}_i^2,\cdots,{x}_i^{T-1}\}}$, GRU updates the cell vector sequence $\small{{s}_i=\{{s}_i^1,{s}_i^2,\cdots,{s}_i^{T-1}\}}$ 
and company hidden state $\small{{u}_i=\{{u}_i^2,{u}_i^3,\cdots,{u}_i^{T}\}}$ from $t=1$ to $t=T-1$. After the initialization, in year $t$, the company state ${u}_i^{t+1}$ is updated by the previous hidden state ${u}_i^{t}$ and the current internal embedding vector ${x}_i^t$, which is shown as:
\begin{equation}
\small 
\label{equ:gru}
\begin{array}{l}
{\boldsymbol{z}_i^{t+1} =\sigma\left({W}_{x z} {x}_i^{t+1}+{W}_{uz} {u}_i^{t}\right)} \\ 
{{r}_i^{t+1} =\sigma\left({W}_{x r} {x}_i^{t+1}+{W}_{u r} {u}_i^{t}\right)} \\ 
{\tilde{{u}}_i^{t+1} =\tanh \left({W}_{x u} {x}_i^{t+1}+{r}_i^{t+1} \odot\left({W}_{u u} {u}_i^{t}\right)\right)} \\ 
{{u}_i^{t+1} =\left(\mathbf{1}-{z}_i^{t+1}\right) \odot \tilde{{u}}_i^{t+1}+{z}_i^{t+1} \odot {u}_i^{t}}\end{array}, \\
\end{equation}
\noindent where ${z}_i^{t+1}$, ${r}_i^{t+1}$ are the update and reset gate, respectively. $\odot$ is an element-wise multiplication and $\sigma$ is non-linear activation function which is stated as \textit{sigmoid} in this paper. ${W}_*$ denotes weight matrices, which are all optimized in training process. In this way, the whole evolving process of $\small U_i$ in year $t$ are embedded into a hidden embedding state ${u}_i^{t+1}$, in different years integrated by different relation-enhanced internal embeddings. 

Similar operations are done for mining dynamics of technologies. Then, the final latent embedding ${v}_j^{t+1}$ of $\small V_j$ in year $t$ is also captured automatically, in different years referring to different relation-enhanced external embedding.

\subsubsection{\textbf{Technology Distribution Forecasting}} 
After the above modules, we acquire the latent embeddings of companies and technologies from year $1$ to $T$, denoted by $\small{u^t=[u_1^t,u_2^t,\cdots,u_M^t]}$ and $\small{v^t=[v_1^t,v_2^t,\cdots,v_N^t]}$. Then, when making predictions, we feed ${u}$ and ${v}$ into a function, $\small \hat{r}_{u v}=\mathcal{P}({u}, {v})$, where $\mathcal{P}$ is an arbitrary prediction function or a prediction neural network. For the sake of simplicity, we set $\hat{r}_{u v}=\sigma ({u}\cdot {v})$, which is more efficient for training and easier to avoid overfitting, and $\sigma$ is a sigmoid function. 

Specially, we adopt the idea of Bayesian Personalized Ranking (BPR)~\cite{rendle2009bpr} for pair-wise learning, which has been widely used in recommendation tasks: 
\begin{equation}
\label{equ:loss}
\small L=\sum_{(i, j) \in \mathcal{D}_{S}} -\ln\sigma(\hat{r}_{ij^+}-\hat{r}_{ij^-}) + \lambda||\Theta||^2,
\end{equation}

\noindent where $\Theta$ includes all model parameters, and $\lambda$ and is the regularization factor. $\mathcal{D}_{S}$ indicates the whole training set, which consists of many triples in form of $(i,j^+,j^-)$, meaning that company $i$ shows a greater emphasis on technology $j^+$ than $j^-$. In order to minimize the above object function, we adopt Adadelta optimizer~\cite{zeiler2012adadelta} to update the model parameters with back propagation algorithm, which can be implemented automatically through Tensorflow\footnote{https://www.tensorflow.org}.
\section{Experiment}
\label{s:experiment}

In this section, extensive experiments are conducted on USPTO patent dataset to verify the effectiveness of Deep Technology Forecasting framework. 

\subsection{Experimental Settings}
The USPTO dataset includes 6,014,932 granted US patents from 1972 to 2017, belonging to 389,246 patent assignees.  
After cleaning, we totally get 2,791 high-tech companies, who have filed at least 200 patents since 1972. In addition, all experiments are conducted based on CPC group, meaning that we aim to make predictions on 662 pre-defined technologies.

For better proving the effectiveness of DTF framework, we divide the patent dataset from 1995 to 2015 into four periods, on which experiments are made separately. Let's take 1995 to 2000 as an example. In training stage, we apply patent filing histories of companies and technologies from 1995 to 1999 as input, and technology distribution in 2000 as a ground truth. For testing, one year is  shifted backwards, i.e. with data from 1996 to 2000 as input and 2011 as the prediction target. Treating it as a ranking problem, we evaluate the performance of DTF by the Normalized Discounted Cumulative Gain ($NDCG@K$, $K=10,20,50,100$). All experiments are implemented on a Linux server with four 2.0GHz Intel Xeon E5-2620 CPUs and a Tesla K20m GPU.

\begin{figure*}[t]
	\centering\includegraphics[height = 1.1in]{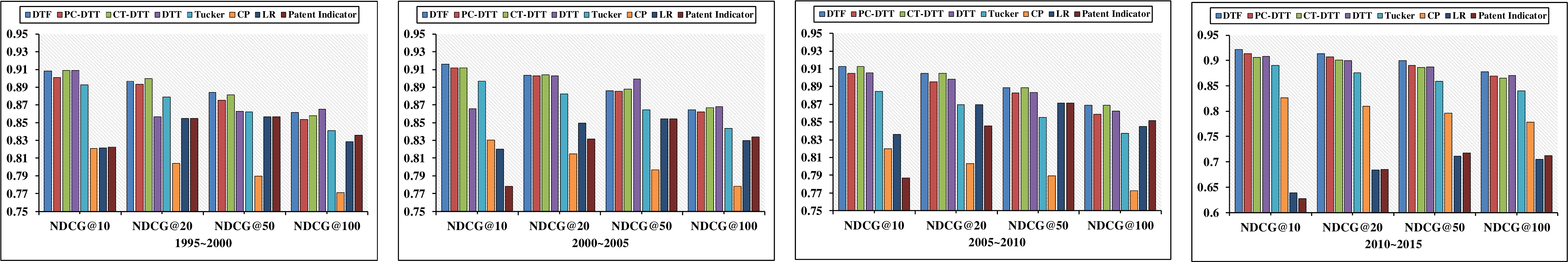}
	\caption{The experimental results on four datasets.} \label{fig:ndcg}
\end{figure*}
\begin{table*}
	\begin{scriptsize}
		\begin{floatrow}
			{\caption{A Case  study  on  Hughes  Net-work  Systems,  LLC  } \label{tab:case}}
			\begin{tabular}{l|c|c|c|c|c|c|c|c|c|c}
				\hline
				\textbf{Methods}         &\multicolumn{10}{c}{\textbf{Predicted Top@10 Technologies}}   \\ \hline
				Ground Truth    &H04L 	&H04W	&H04B	&H03M	&H03H	&G06F	&H04M	&H01Q	&G06E	&A44B   \\ \hline
				DTF             &H04L	&H04W	&H04B	&H03M	&H04M	&G10C	&H04Q	&B60G	&Y02W	&G09F   \\ \hline
				PC-DTT          &H04L	&H04W	&H04B	&H03M	&C12Q	&G10D	&B60G	&F02C	&F16M	&H04M   \\ \hline
				CT-DTT          &H04L	&H04W	&H04B	&H03M	&G06F	&D02H	&G06C	&E21B	&H04Q	&C12P   \\ \hline
				DTT             &H04L	&H04W	&H04B	&H03M	&C23F	&D06C	&Y10T	&C22B	&H04J	&F42D   \\ \hline
			\end{tabular}    
			\begin{tabular}{c|c}
				\hline
				\textbf{Codes}  &\textbf{Meanings} \\ \hline
				H04L    & Transmission of digital information \\ \hline
				H04W    & Wireless communication networks\\ \hline
				H04B    & Transmission systems \\ \hline
				H03M    & Coding; Decoding; Code conversion \\ \hline
				H03H    & Impedance networks\\ \hline
			\end{tabular} \label{tab:group}  
		\end{floatrow}
	\end{scriptsize}
	\vspace{-4mm}
\end{table*}

\subsection{Compared Methods}
Since there are few prior works to directly predict the possible technologies customized to companies' personalized R\&D needs, we introduce some variants of DTF to highlight the effectiveness of each component of our framework. 
\begin{itemize}[leftmargin=*]
    \item \textit{PC-DTT} excludes the collaborative relations among technologies as the input of DTT.
    \item \textit{CT-DTT} excludes the competitve relations among technologies as the input of DTT.
	\item \textit{DTT} only inputs the patent filing history of companies and technologies as well as their dynamic interactions.
	\item \textit{CP}~\cite{kolda2009tensor} only models the dynamic interactions between companies and technologies. 
	\item \textit{Tucker}~\cite{de2000multilinear} has the same settings with \textit{CP}.
	\item \textit{LR} ignores the dynamic embeddings of companies and technologies.
	\item \textit{Patent Indicator}~\cite{kyebambe2017forecasting} can also give useful advice for predicting emerging technologies in special technology fields.
	
\end{itemize}

\subsection{Experimental Results}
Fig.~\ref{fig:ndcg} shows the performances of DTF and compared methods within four time periods. Here, we can observe that in most cases DTF performs much better than baselines under all metrics with respect to different $K$, indicating that it is meaningful to integrate both the relation-enhanced internal and external factors along with dynamic interactions among companies and technologies. 

Among DTF and its variants, DTF often performs best, proving the effectiveness of competitions extracted by PCR and collaborations extracted by CTR. What's more, there seems a tight race between PC-DTT and CT-DTT: on the first three datasets, CT-DTT shows a great advantage beyond PC-DTT, while on the last one, CT-DTT behaves much better than CT-DTT. This phenomenon may indicate that competitive relations among companies have gradually become more and more important for technology tracing. 

Compared with baselines including Tucker, CP and LR, DTF still behaves better. For one thing, although Tucker and CP model the same dynamic interactions, they yet do not perform very well, which proves that patent content information can be very useful for mining technology distribution. For another, LR integrates the yearly content information the same as DTF while shows a bed performance, especially when $K$ is set as 10 and 20, indicating the fact that dynamic interactions among companies and technologies can not be ignored.

In the end, almost all models behave better from 1999 to 2010 except for Patent Indicator, which is understandable in that patents filed in recent years haven't received many citations, so statistics-based Patent Indicators have no access to distinctive features 
(especially citation-based features). However, DTF shows an advantage in this term, because it tries to learn potential semantic information from many patent documents, depending less on statistics-based features.

\subsection{Case Study.} 

In this section, we present a case study on Hughes Network Systems, LLC (Hughes), which is the global leader in broadband satellite technology and services for home and office\footnote{https://www.hughes.com}. Table~\ref{tab:case} shows top 10 technologies in 2016 of Hughes predicted by DTF and its variants. From this table, we can see that both DTF and its variants successfuly predict LLC (Hughes) will pay the most emphasis on technologies about network communication, represented by CPC codes as 'H04L', 'H04W', 'H04B', and 'H03M'. However, about the followings, they have very different ideas: 1) Both DTF and PC-DTT prefer 'B60G' (Vehicle suspension arrangements), which may give a signal that its competitors may have some businesses in this field;
2) Both DTF and CT-DTT think 'H04Q' (switches, relays etc.) will be an important technology for Hughes, which might be due to the big collaboration degrees with the former technologies, especially 'H04W'. In fact, they share 39,898 common patents according to our statistics. 
\section{Related Work}
\label{s:related-work}
Patent data has been widely explored for decision-making processes and strategic planning purposes~\cite{ernst2003patent, kim2017novel, erdi2013prediction}. Typically, methods related to technology prospecting can be summarized as two types: qualitative analysis and quantitative mining. Qualitative approaches are mainly based on analysis by domain experts, which naturally needs many human efforts, and in addition, some researches~\cite{lee2017technology} find that these subjective strategies may be not always precisely correct and reliable. Quantified approaches aim to access potential prospects of technologies through supervised machine learning methods~\cite{kyebambe2017forecasting, erdi2013prediction}. 

Nowadays, deep learning has been widely used in many traditional areas, i.e. education~\cite{huang2019ekt}, financial analyses~\cite{zhang2018caden}, music generation~\cite{zhu2018xiaoice}, patent mining~\cite{liu2018patent}, and etc.
In particular, Recurrent Neural Networks (RNN) are powerful tools for modeling sequences, which are flexibly extensible and can incorporate various kinds of information including temporal order~\cite{donkers2017sequential}. 
Its variants, such as Long Short-Term Memory~(LSTM)~\cite{graves2013speech} and Gated Recurrent Unit~(GRU)~\cite{cho2014properties}, have capability to model dependency among sequences.

\section{Conclusion and Future Work}
\label{s:conclusion}
In this paper, we presented a focused study on technology tracing problem. Specifically, we designed a novel data-driven Deep Technology Forecasting (DTF) framework including three components: Potential Competitor Recognition (PCR), Collaborative Technology Recognition (CTR), and Deep Technology Tracing (DTT) neural network. For one thing, PCR aimed to capture the competitive relations among enterprises and CTR tried to figure out the collaborative relations among technologies. For another, DTT targeted at modeling dynamic interactions between companies and technologies. Finally, we evaluated our DTF framework on real-world patent data and the experimental results clearly proved its effectiveness. 
We hope this work could lead to more future studies.

\section{Acknowledgements}
This research was partially supported by grants from the National Key Research and Development Program of China (No. 2018YFC0832101), the National Natural Science Foundation of China (Grants No., 61672483, 61727809), the Young Elite Scientist Sponsorship Program of CAST and the Youth Innovation Promotion Association of CAS (No. 2014299).

\small
\bibliographystyle{unsrt}
\bibliography{ICDM_2019}

\begin{thebibliography}{10}

\bibitem{kurtossy2004innovation}
Jen{\H{o}} K{\"u}rt{\"o}ssy.
\newblock Innovation indicators derived from patent data.
\newblock {\em Periodica Polytechnica Social and Management Sciences},
  12(1):91--101, 2004.

\bibitem{park2017application}
Youngjin Park and Janghyeok Yoon.
\newblock Application technology opportunity discovery from technology
  portfolios: Use of patent classification and collaborative filtering.
\newblock {\em Technological Forecasting and Social Change}, 118:170--183,
  2017.

\bibitem{ernst2003patent}
Holger Ernst.
\newblock Patent information for strategic technology management.
\newblock {\em World patent information}, 25(3):233--242, 2003.

\bibitem{kim2017novel}
Gabjo Kim and Jinwoo Bae.
\newblock A novel approach to forecast promising technology through patent
  analysis.
\newblock {\em TFSC}, 117:228--237, 2017.

\bibitem{erdi2013prediction}
P{\'e}ter {\'E}rdi, Kinga Makovi, Zolt{\'a}n Somogyv{\'a}ri, Katherine
  Strandburg, Jan Tobochnik, P{\'e}ter Volf, and L{\'a}szl{\'o} Zal{\'a}nyi.
\newblock Prediction of emerging technologies based on analysis of the us
  patent citation network.
\newblock {\em Scientometrics}, 95(1):225--242, 2013.

\bibitem{del1999resource}
Jesus~Galende Del~Canto and Isabel~Suarez Gonzalez.
\newblock A resource-based analysis of the factors determining a firm's r\&d
  activities.
\newblock {\em Research Policy}, 28(8):891--905, 1999.

\bibitem{liu2018patent}
Qi~Liu, Han Wu, Yuyang Ye, Hongke Zhao, Chuanren Liu, and Dongfang Du.
\newblock Patent litigation prediction: A convolutional tensor factorization
  approach.
\newblock In {\em IJCAI}, pages 5052--5059, 2018.

\bibitem{goldberg2016primer}
Yoav Goldberg.
\newblock A primer on neural network models for natural language processing.
\newblock {\em J. Artif. Intell. Res.(JAIR)}, 57:345--420, 2016.

\bibitem{cho2014properties}
Kyunghyun Cho, Bart Van~Merri{\"e}nboer, Dzmitry Bahdanau, and Yoshua Bengio.
\newblock On the properties of neural machine translation: Encoder-decoder
  approaches.
\newblock {\em arXiv preprint arXiv:1409.1259}, 2014.

\bibitem{rendle2009bpr}
Steffen Rendle, Christoph Freudenthaler, Zeno Gantner, and Lars Schmidt-Thieme.
\newblock Bpr: Bayesian personalized ranking from implicit feedback.
\newblock In {\em UAI}, pages 452--461. AUAI Press, 2009.

\bibitem{zeiler2012adadelta}
Matthew~D Zeiler.
\newblock Adadelta: an adaptive learning rate method.
\newblock {\em arXiv preprint arXiv:1212.5701}, 2012.

\bibitem{kolda2009tensor}
Tamara~G Kolda and Brett~W Bader.
\newblock Tensor decompositions and applications.
\newblock {\em SIAM review}, 51(3):455--500, 2009.

\bibitem{de2000multilinear}
Lieven De~Lathauwer, Bart De~Moor, and Joos Vandewalle.
\newblock A multilinear singular value decomposition.
\newblock {\em SIAM journal on Matrix Analysis and Applications},
  21(4):1253--1278, 2000.

\bibitem{kyebambe2017forecasting}
Moses~Ntanda Kyebambe, Ge~Cheng, Yunqing Huang, Chunhui He, and Zhenyu Zhang.
\newblock Forecasting emerging technologies: A supervised learning approach
  through patent analysis.
\newblock {\em Technological Forecasting and Social Change}, 125:236--244,
  2017.

\bibitem{lee2017technology}
Jeongjin Lee, Changseok Kim, and Juneseuk Shin.
\newblock Technology opportunity discovery to r\&d planning: Key technological
  performance analysis.
\newblock {\em Technological Forecasting and Social Change}, 119:53--63, 2017.

\bibitem{huang2019ekt}
Zhenya Huang, Yu~Yin, Enhong Chen, Hui Xiong, Yu~Su, Guoping Hu, et~al.
\newblock Ekt: Exercise-aware knowledge tracing for student performance
  prediction.
\newblock {\em IEEE TKDE}, 2019.

\bibitem{zhang2018caden}
Liang Zhang, Keli Xiao, Hengshu Zhu, Chuanren Liu, Jingyuan Yang, and Bo~Jin.
\newblock Caden: A context-aware deep embedding network for financial opinions
  mining.
\newblock In {\em IEEE ICDM}, pages 757--766. IEEE, 2018.

\bibitem{zhu2018xiaoice}
Hongyuan Zhu, Qi~Liu, Nicholas~Jing Yuan, Chuan Qin, Jiawei Li, Kun Zhang,
  Guang Zhou, Furu Wei, Yuanchun Xu, and Enhong Chen.
\newblock Xiaoice band: A melody and arrangement generation framework for pop
  music.
\newblock In {\em SIGKDD}, pages 2837--2846. ACM, 2018.

\bibitem{donkers2017sequential}
Tim Donkers, Benedikt Loepp, and J{\"u}rgen Ziegler.
\newblock Sequential user-based recurrent neural network recommendations.
\newblock In {\em Proceedings of the Eleventh ACM Conference on Recommender
  Systems}, pages 152--160. ACM, 2017.

\bibitem{graves2013speech}
Alex Graves, Abdel-rahman Mohamed, and Geoffrey Hinton.
\newblock Speech recognition with deep recurrent neural networks.
\newblock In {\em 2013 IEEE international conference on acoustics, speech and
  signal processing}, pages 6645--6649. IEEE, 2013.

\end{thebibliography}

\end{document}